\documentclass[runningheads]{llncs}
\usepackage[utf8]{inputenc}
\usepackage[T1]{fontenc}
\usepackage{graphicx}
\usepackage[space]{grffile}
\usepackage{hyperref}

\usepackage[french,USenglish]{babel}
\usepackage{xspace}
\usepackage{amsmath}
\usepackage{amsfonts}

\DeclareMathOperator{\R}{\mathbb{R}}

\DeclareMathOperator{\N}{\mathbb{N}}

\makeatletter
\DeclareRobustCommand\onedot{\futurelet\@let@token\@onedot}
\def\@onedot{\ifx\@let@token.\else.\null\fi\xspace}

\def\eg{\emph{e.g}\onedot} 
\def\ie{\emph{i.e}\onedot} 
\def\cf{\emph{cf.}\xspace} 

 \def\vs{\emph{vs}\onedot}
 
\def\etal{\emph{et~al}\onedot}
\makeatother

\newcommand{\datahela}{\mbox{\textsc{HeLa}}\xspace}
\newcommand{\dataalan}{\mbox{\textsc{Mdck}}\xspace}
\newcommand{\dataflywing}{\mbox{\textsc{Flywing}}\xspace}
\newcommand{\datayeast}{\mbox{\textsc{Yeast}}\xspace}
\newcommand{\tap}{\mbox{\textsc{Tap}}\xspace}
\newcommand{\unet}{\mbox{\textsc{U-Net}}\xspace}
 \usepackage[usenames,dvipsnames]{xcolor}

\usepackage{palatino}
\usepackage{geometry}
\hypersetup{colorlinks = true,
            linkcolor = black,
            urlcolor  = black,
            citecolor = black}
\usepackage{makecell}
\usepackage{booktabs}

\usepackage
{todonotes}

\usepackage[capitalise]{cleveref}

\usepackage{float}

\usepackage[font=small,skip=10pt,labelfont=bf]{caption}

\usepackage{siunitx}

\begin{document}
\title{Self-supervised dense representation learning for live-cell microscopy with time arrow prediction}
\titlerunning{Self-supervised dense representation learning for live-cell microscopy}
\author{Benjamin Gallusser \and Max Stieber \and Martin Weigert}
\authorrunning{Gallusser et al.}
\institute{École polytechnique fédérale de Lausanne (EPFL)
\href{mailto:martin.weigert@epfl.ch}{\tt \{benjamin.gallusser,max.stieber,martin.weigert\}@epfl.ch}\\
}
\maketitle

\begin{abstract}
State-of-the-art object detection and segmentation methods for microscopy images rely on supervised machine learning, which requires laborious manual annotation of training data.
Here we present a self-supervised method based on \emph{time arrow prediction pre-training} that learns dense image representations from raw, unlabeled live-cell microscopy videos.
Our method builds upon the task of predicting the correct order of time-flipped image regions via a single-image feature extractor followed by a time arrow prediction head that operates on the fused features.
We show that the resulting dense representations capture inherently time-asymmetric biological processes such as cell divisions on a pixel-level. 
We furthermore demonstrate the utility of these representations on several live-cell microscopy datasets for detection and segmentation of dividing cells, as well as for cell state classification. Our method outperforms supervised methods, particularly when only limited  ground truth annotations are available as is commonly the case in practice. We provide code at \url{https://github.com/weigertlab/tarrow}.

\end{abstract}

\section{Introduction}

Live-cell microscopy is a fundamental tool to study the spatio-temporal dynamics of biological systems~\cite{tomer_shedding_2011,etournay_interplay_2015,stelzer_light_2021}.
The resulting datasets can consist of terabytes of raw videos that require automatic methods for downstream tasks such as classification, segmentation, and tracking of objects (\eg cells or nuclei).
Current state-of-the-art methods rely on supervised learning using deep neural networks
that are trained on large amounts of ground truth annotations~\cite{weigert2020,stringer2021,greenwald_whole-cell_2021}.
The manual creation of these annotations, however, is laborious and often constitutes a practical bottleneck in the analysis of microscopy experiments~\cite{greenwald_whole-cell_2021}.
Recently, self-supervised representation learning (SSL) has emerged as a promising approach to alleviate this problem~\cite{ericsson_self-supervised_2022,chen2020simple}.
In SSL one first defines a \emph{pretext task} which can be formulated solely based on \emph{unlabeled} images (\eg inpainting~\cite{he2022masked}, or rotation prediction~\cite{gidaris_unsupervised_2018}) and tasks a neural network to solve it, with the aim of generating latent representations that capture high-level image semantics.
In a second step, these representations can then be either \emph{finetuned} or used directly (\eg via \emph{linear probing}) for a \emph{downstream task} (\eg image classification) with available ground truth~\cite{pathak_context_2016,hsu2021,han2022}.
Importantly, a proper choice of the pretext task is crucial for the resulting representations to be beneficial for a specific downstream task.
\begin{figure}[t]
  \includegraphics[width=1.0\textwidth]{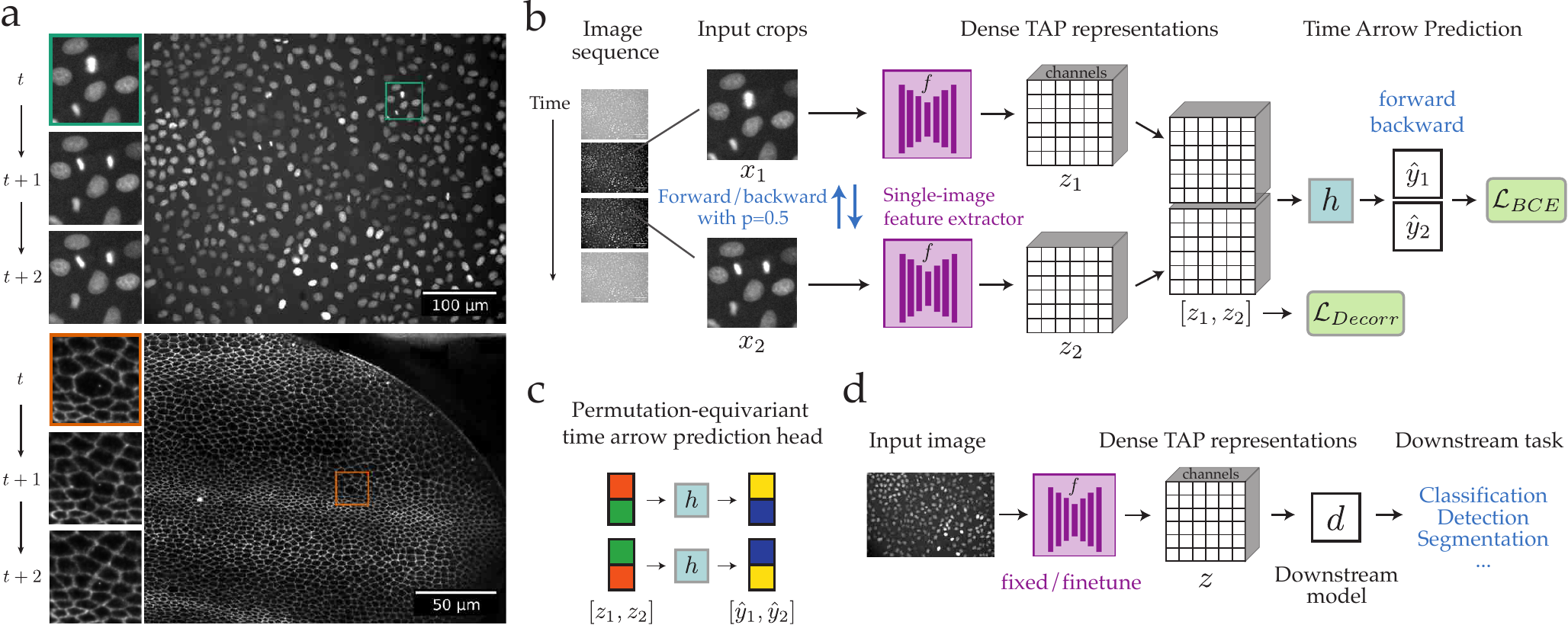}
	\caption{%
		\textbf{a)} Example frames from two live-cell microscopy videos. Top: \emph{MDCK} cells with labeled nuclei~\cite{ulicna_automated_2021}, Bottom: \emph{Drosophila} wing with labeled membrane~\cite{etournay_interplay_2015}. Insets show three consecutive time points containing cell divisions. 
		\textbf{b)} Overview of \tap: We create crops $(x_1,x_2)$ from consecutive time points of a given video. After randomly flipping the input order (forward/backward), each crop is passed through a dense feature extractor $f$ creating pixel-wise \tap representations $(z_1,z_2)$. These are stacked and fed to the time arrow prediction head $h$.
		\textbf{c)} We design $h$ to be permutation-equivariant ensuring consistent classification of temporally flipped inputs.
		\textbf{d)} The learned \tap representations $z$ are used as input to a downstream model $d$.
	}
	\label{fig:overview}
  \end{figure}
  
In this paper we investigate whether \emph{time arrow prediction}, \ie the prediction of the correct order of temporally shuffled image frames extracted from live-cell microscopy videos, can serve as a suitable pretext task to generate meaningful representations of microscopy images.
We are motivated by the observation that for most biological systems the temporal dynamics of local image features are closely related to their semantic content: whereas static background regions are time-symmetric, processes such as cell divisions or cell death are inherently time-asymmetric (\cf~\cref{fig:overview}a).
Importantly, we are interested in \emph{dense} representations of individual images as they are useful for both image-level (\eg classification) or pixel-level (\eg segmentation) downstream tasks.
To that end, we propose  a time arrow prediction pre-training scheme, which we call \tap,  that uses a feature extractor operating on single images followed by a time arrow prediction head operating on the fused representations of consecutive time points.
The use of time arrow prediction as a pretext task for natural (\eg youtube) videos was introduced by Pickup~\etal~\cite{pickup_seeing_2014} and has since then seen numerous applications for image-level tasks, such as action recognition, video retrieval, and motion classification~\cite{misra_shuffle_2016,lee_unsupervised_2017,wei_learning_2018,schiappa_self-supervised_2022,dorkenwald_scvrl_2022,hu_contrast_2021}.
However, to the best of our knowledge, SSL via time arrow prediction has not yet been studied in the context of live-cell microscopy.
Concretely our contributions are: 
\emph{i)} We introduce the time arrow prediction pretext task to the domain of live-cell microscopy and propose the \tap pre-training scheme, which learns dense representations (in contrast to only image-level representations) from raw, unlabeled live-cell microscopy videos, 
\emph{ii)} we propose a custom (permutation-equivariant) time arrow prediction head that enables robust training,
\emph{iii)} we show via attribution maps that the representations learned by \tap capture biologically relevant processes such as cell divisions, and finally
\emph{iv)} we demonstrate that \tap representations are beneficial for common image-level and pixel-level downstream tasks in live-cell microscopy, especially in the low training data regime.

\section{Method}

Our proposed \tap pre-training takes as input a set $\{I\}$ of live-cell microscopy image sequences $I \in \R^{T \times H \times W}$ with the goal to produce a feature extractor $f$ that generates $c$-dimensional dense representations $z = f(x) \in \R^{c \times H \times W}$ from single images $x \in R^{H \times W}$  (\cf~\cref{fig:overview}b for an overview of \tap). To that end, we randomly sample from each sequence $I$ pairs of smaller patches $x_1, x_2 \in \R^{h \times w}$ from the same spatial location but consecutive time points $x_1 \subset I_t, x_2 \subset I_{t+1}$.
We next flip the order of each pair with equal probability $p=0.5$, assign it the corresponding label $y$ (\textit{forward} or \textit{backward}) and compute dense representations $z_1=f(x_1)$ and $z_2=f(x_2)$ with $z_1, z_2 \in \R^{c \times h \times w}$ via a fully convolutional feature extractor $f$.
The stacked representations $z = [z_1, z_2] \in \R^{2 \times c \times h \times w}$ are fed to a \emph{time arrow prediction head} $h$, which produces the classification logits $\hat{y} = [\hat{y}_1,\hat{y}_2] = h([z_1, z_2]) = h([f(x_1),f(x_2)])  \in \R^2$.
Both $f$ and $h$ are trained jointly to minimize the loss
\begin{align}
\mathcal{L} = \mathcal{L}_{BCE}(y,\hat{y}) + \lambda \mathcal{L}_{Decorr}(z)\ ,
\end{align}
where $ \mathcal{L}_{BCE}$ denotes the standard softmax + binary cross-entropy loss between the ground truth label $y$ and the logits $\hat{y} = h(z)$, and  $\mathcal{L}_{Decorr}$ is a loss term that promotes $z$ to be decorrelated across feature channels~\cite{zbontar_barlow_2021,hua_feature_2021} via maximizing the diagonal of the softmax-normalized correlation matrix $A_{ij}$:
\begin{align}
\mathcal{L}_{Decorr}(\tilde{z}) = - \frac{1}{c} \text{log} \sum_{i=1}^c A_{ii}\ , \quad A_{ij} = \mathrm{softmax}(\tilde{z}_i^T \! \cdot \tilde{z}_j / \tau) =  \frac{e^{\tilde{z}_i^T \! \cdot \tilde{z}_j / \tau}}{\sum_{j=1}^c e^{\tilde{z}_i^T \! \cdot \tilde{z}_j / \tau}}
\end{align}
Here $\tilde{z} \in \R^{c \times 2hw}$ denotes the stacked features $z$ flattened across the non-channel dimensions, and $\tau$ is a temperature parameter.
Throughout the experiments we use $\lambda = 0.01$ and $\tau=0.2$.
Note that instead of creating image pairs from consecutive video frames we can as well choose a custom time step $\Delta t \in \N$ and sample  $x_1 \subset I_t$ and $x_2 \subset I_{t+\Delta t}$, which we empirically found to work better for datasets with high frame rate. 
\subsubsection*{Permutation-equivariant time arrow prediction head:}
 The time arrow prediction task has an inherent symmetry: flipping the input $[z_1, z_2] \to [z_2, z_1]$  should flip the logits  $[\hat{y}_1, \hat{y}_2] \to [\hat{y}_2, \hat{y}_1]$. In other words, $h$ should be \emph{equivariant} wrt.\ to permutations of the input. In contrast to common models (\eg ResNet~\cite{he2016deep}) that lack this symmetry, we here directly incorporate this inductive bias via a \emph{permutation-equivariant head} $h$ that is a generalization of the set permutation-equivariant layer proposed in~\cite{zaheer_deep_2017} to dense inputs.
 Specifically, we choose $h = h_1 \circ  \ldots \circ  h_L $ as a chain of permutation-equivariant layers $h_l$:
\begin{align}
h_l & : \R^{2 \times c \times h \times w} \to \R^{2 \times \tilde{c} \times h \times w} \nonumber \\
h_l(z)_{tmij} &= \sigma \big( \sum_{n} L_{mn} z_{t,n,i,j} +  \sum_{s,n} G_{mn} z_{s,n,i,j} \big)\ ,
\end{align}
with weight matrices $L,G \in \R^{\tilde{c} \times c}$ and a non-linear activation function $\sigma$. 
Note that $L$ operates independently on each temporal axis and thus is trivially permutation equivariant, while $G$ operates on the temporal sum and thus is permutation invariant. The last layer $h_L$ includes an additional global average pooling along the spatial dimensions to yield the final logits $\hat{y} \in \R^2$.

\subsubsection*{Augmentations:}
To avoid overfitting on artificial image cues that could be discriminative of the temporal order (such as a globally consistent cell drift, or decay of image intensity due to photo-bleaching) we apply the following augmentations (with probability 0.5) to each image patch pair $x_1, x_2$:
flips, arbitrary rotations and elastic transformations (jointly for $x_1$ and $x_2$), translations for $x_1$ and $x_2$ (independently), spatial scaling, additive Gaussian noise, and intensity shifting and scaling (jointly+independently).

\section{Experiments}
\subsection{Datasets}
\begin{figure}[t]
	\includegraphics[width=1.0\textwidth]{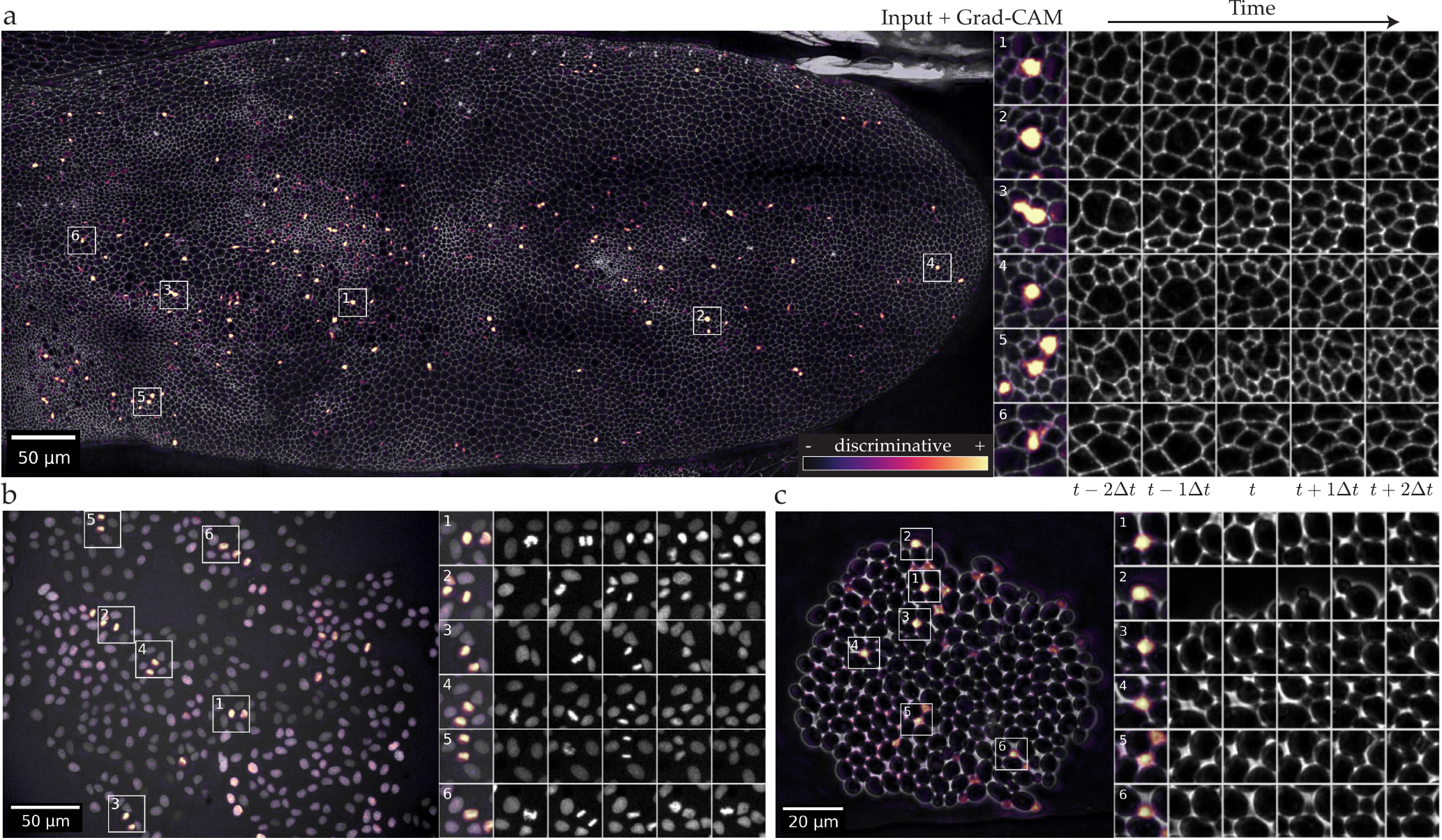}
  \caption{A single image frame overlayed with \tap attribution maps (computed with Grad-CAM~\cite{selvaraju_grad-cam_2017}) for \textbf{a}) \dataflywing, \textbf{b}) \dataalan, and \textbf{c}) \datayeast. Insets show the top six most discriminative regions and their temporal context ($\pm$ 2 timepoints). Note that across all datasets almost all regions  contain cell divisions. Best viewed on screen.}
	\label{fig:cams}
\end{figure}

To demonstrate the utility of \tap for a diverse set of specimen and microscopy modalities we use the following four different datasets:
\vspace{0.15cm}
\\
\textbf{\datahela}
Human cervical cancer cells expressing histone 2B–GFP imaged by fluorescence microscopy every 30 minutes~\cite{ulman_objective_2017} . The dataset consists of four videos with overall 368 frames of size $1100 \times 700$. We use $\Delta t=1$ for \tap training.\\
\textbf{\dataalan}
Madin-Darby canine kidney epithelial cells expressing histone 2B–GFP (\cf~\cref{fig:cams}b), imaged by fluorescence microscopy every 4 minutes~\cite{ulicna_automated_2021,mdck_download_link}. The dataset consists of a single video with 1200 frames of size $1600 \times 1200$. We use $\Delta t \in \{4,8\}$.\\
\textbf{\dataflywing}
 \textit{Drosphila melanogaster} pupal wing expressing Ecad::GFP (\cf~\cref{fig:cams}a), imaged by spinning disk confocal microscopy every 5 minutes~\cite{piscitello-gomez_core_2022,etournay_interplay_2015}. The dataset consists of three videos with overall 410 frames of size $3900 \times 1900$. We use $\Delta t=1$.\\
\textbf{\datayeast}
\textit{S. cerevisiae} cells (\cf~\cref{fig:cams}c) imaged by phase-contrast microscopy every 3 minutes~\cite{padovani_segmentation_2022,padovani_cell-acdc_2022}. The dataset consists of five videos  with overall 600 frames of size $1024 \times 1024$. We use $\Delta t \in \{1,2,3\}$.
\vspace{0.15cm}
\\
For each dataset we heuristically choose $\Delta t$ to roughly correspond to the time scale of observable biological processes (\ie larger $\Delta t$ for higher frame rates).

\begin{figure}[t]
  \includegraphics[width=1.0\textwidth]{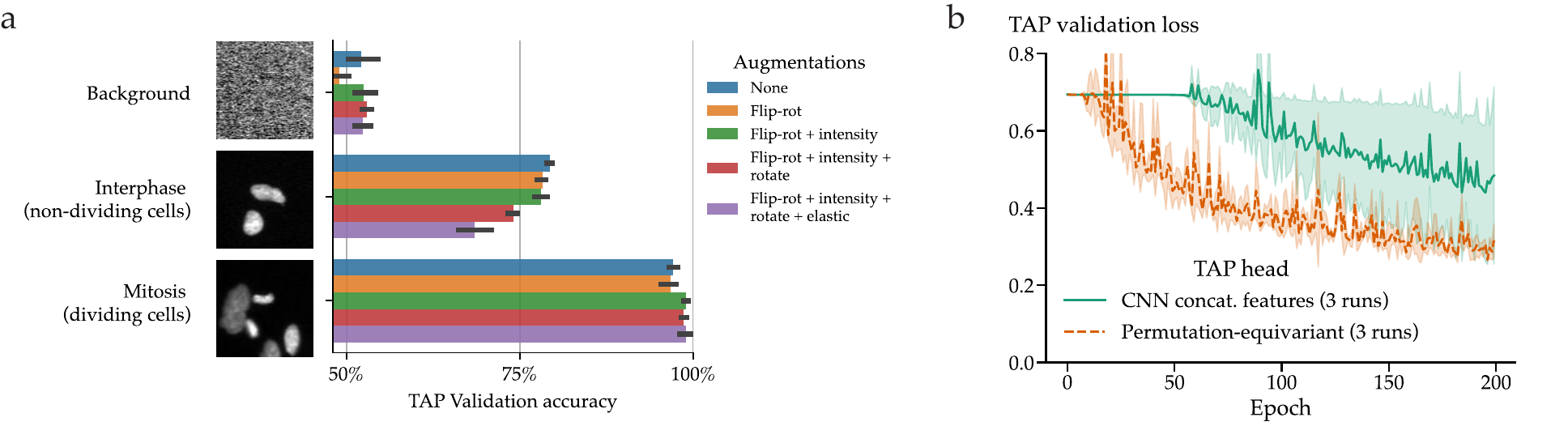}
	\caption[caption]{%
		\textbf{a)} \tap validation accuracy for different image augmentations on crops of background, interphase (non-dividing), and mitotic (dividing) cells (from \datahela dataset).
		\textbf{b)} \tap validation loss during training on \dataflywing for a regular CNN time arrow prediction head (green) and the proposed permutation-equivariant head (orange). We show results of three runs per model.   
	}
	\label{fig:taptraining}
\end{figure}

\subsection{Implementation details:}
For the feature extractor $f$ we use a 2D \unet~\cite{ronneberger_u-net_2015} with depth 3 and $c=32$ output features, batch normalization and leaky ReLU activation (approx.~2M params). The time arrow prediction head $h$ consists of two per\-mu\-ta\-tion-equivariant layers with batch normalization and leaky ReLU activation, followed by global average pooling and a final per\-mu\-ta\-tion-equivariant layer (approx.~5k params). We train all \tap models for 200 epochs and $10^5$ samples per epoch, using the Adam optimizer~\cite{kingma2015} with a learning rate of $4\times10^{-4}$ with cyclic schedule, and batch size 256. Total training time for a single \tap model is roughly 8h on a single GPU. \tap is implemented in PyTorch.

\subsection{Time arrow prediction pretraining}
We first study how well the time arrow prediction pretext task can be solved depending on  different image structures and used data augmentations.
To that end, we train \tap networks with an increasing number of augmentations on \datahela and compute the \tap classification accuracy for consecutive image patches $x_1,x_2$ that contain either background, interphase (non-dividing) cells, or mitotic (dividing) cells.
As shown in~\cref{fig:taptraining}a, the accuracy on background regions is approx.\ 50\% irrespective of the used augmentations, suggesting the absence of predictive cues in the background for this dataset.  
In contrast, on regions with cell divisions the accuracy reaches almost 100\%, confirming that \tap is able to pick up on strong time-asymmetric image features.
Interestingly, the accuracy for regions with non-dividing cells ranges from 68\% to 80\%, indicating the presence of weak visual cues such as global drift or cell growth. When using more data augmentations the accuracy decreases by roughly 12 percentage points, suggesting that data augmentation is key to avoid overfitting on confounding cues.

Next we investigate which regions in full-sized videos are most discriminative for \tap. To that end, we apply a trained \tap network on consecutive full-sized frames $x_1, x_2$  and compute the dense attribution map of the classification logits $y$ wrt. to the \tap representations $z$ via Grad-CAM~\cite{selvaraju_grad-cam_2017}. In~\cref{fig:cams} we show example attribution maps on top of single raw frames for three different datasets.
Strikingly, the attribution maps highlight only a few distributed, yet highly localized image regions.
When inspecting the top six most discriminative regions and their temporal context for a single image frame, we find that virtually all of them contain cell divisions~(\cf~\cref{fig:cams}).
Moreover, when examining the attribution maps for full videos, we find that indeed most highlighted regions correspond to mitotic cells, underlining the strong potential of \tap to reveal time-asymmetric biological phenomena from raw microscopy videos alone (\cf Supplementary Video 1).

\begin{figure}[t]
  \includegraphics[width=1.0\textwidth]{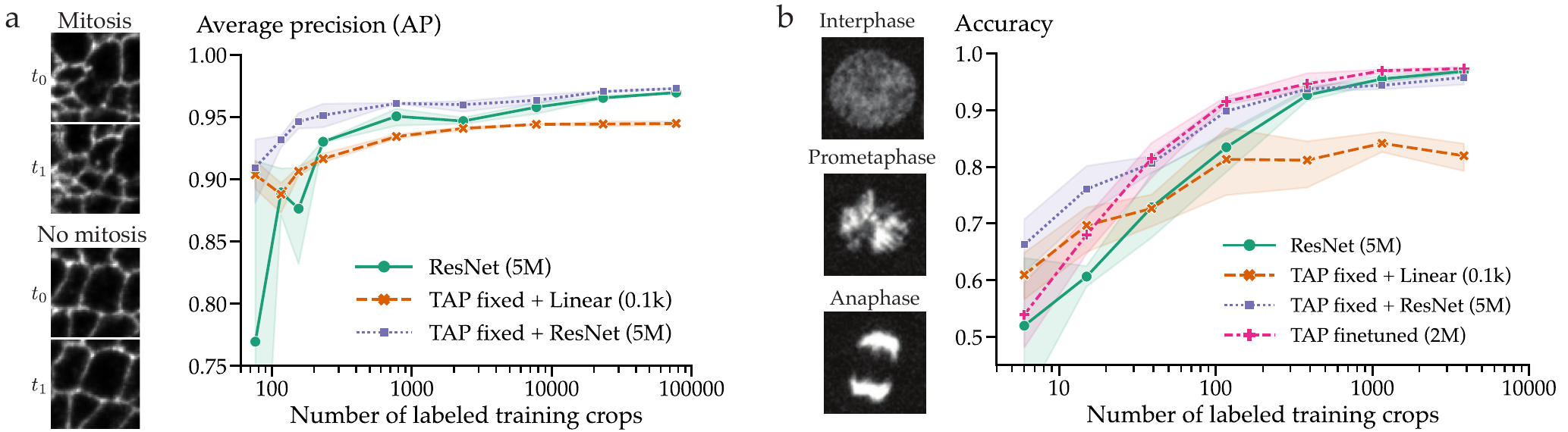}
\caption[caption]{%
	\textbf{a)} Mitosis classification on \dataflywing for two consecutive timepoints with \tap representations \vs a supervised ResNet baseline (green).\hspace{-.9pt}
	\textbf{b)} Cell state classification in \dataalan with fixed/fine-tuned \tap representations \vs a supervised ResNet baseline (green).
	We show results of three runs per model, \# of params in parenthesis.
}
\label{fig:classification}
\end{figure}

Finally, we emphasize the positive effect of the permutation-equivariant time arrow prediction head on the training process.
When we originally used a regular CNN-based head, we consistently observed that the \tap loss stagnated during the initial training epochs and decreased only slowly thereafter~(\cf~\cref{fig:taptraining}b).
Using the permutation-equivariant head alleviated this problem and enabled a consistent loss decrease already from the beginning of training.

\subsection{Downstream tasks}

We next investigate whether the learned \tap representations are useful for common supervised downstream tasks, where we especially focus on their utility in the low training data regime.
First we test the learned representations on two image-level classification tasks, and later on two dense segmentation tasks.
\subsubsection*{Mitosis classification on \dataflywing:}
Since \tap attribution maps strongly highlight cell divisions, we consider predicting mitotic events an appropriate first downstream task to evaluate \tap.
To that end, we generate a dataset of $\num{97}$k crops of size $2 \times 96 \times 96$ from \dataflywing and label them as mitotic/non-mitotic ($\num{16}$k/$\num{81}$k) based on available tracking data~\cite{piscitello-gomez_core_2022}.
We train \tap networks on \dataflywing and use a small ResNet architecture ($\approx 5$M params) that is trained from scratch as a supervised baseline.
In~\cref{fig:classification}a we show average precision (AP) on a held-out test set while varying the amount of available training data. 
As expected, the performance of the supervised baseline drops substantially for low amounts of training data and surprisingly is already outperformed by a linear classifier (100 params) on top of \tap representations (\eg 0.90 \vs 0.77 for 76 labeled crops).
Training a small ResNet on fixed \tap representations consistently outperforms the supervised baseline even if hundreds of annotated cell divisions are available for training (\eg 0.96 \vs 0.95 for 2328 labeled crops with $\sim$ 400 cell divisions), confirming the value of \tap representations to detect mitotic events. 
\subsubsection*{Cell state classification on \dataalan:}
Next we turn to the more challenging task of distinguishing between cells in interphase, prometaphase and anaphase from \dataalan.
This dataset consists of 4800 crops of size $80 \times 80$ that are labeled with one of the three classes (1600 crops/class). Again we use a ResNet as supervised baseline and report in~\cref{fig:classification}b test classification accuracy for varying amount of training data. 
As before, both a linear classifier as well as a ResNet trained on fixed \tap representations outperform the baseline especially in the low data regime, with the latter showing better or comparable results across the whole data regime (\eg 0.90 vs.~0.83 for 117 annotated cells).
Additionally, we finetune the pretrained \tap feature extractor for this downstream task, which slightly improves the results given enough training data.
Notably, already at 30\% training data it reaches the same performance (0.97) as the baseline model trained on the full training set.
\begin{figure}[t]
  \includegraphics[width=1.0\textwidth]{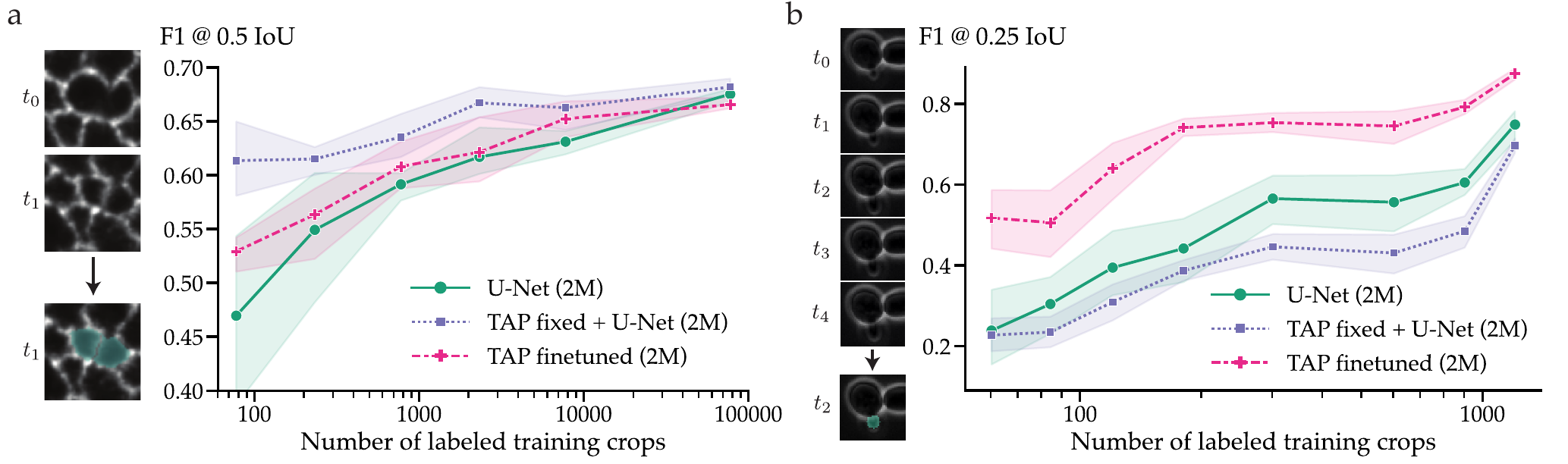}
\caption[caption]{%
	\textbf{a)} Mitosis segmentation in \dataflywing for two consecutive timepoints with fixed/finetuned \tap representations \vs a supervised \unet baseline (green). We report F1 @ 0.5 IoU after removing objects smaller than 64 pixels.
	\textbf{b)} Emerging bud detection in \datayeast from five consecutive timepoints with fixed/finetuned \tap representations versus a supervised \unet baseline (green). We report F1 @ 0.25 IoU on 2D+time objects.
	We show results of three runs per model, \# of params in parenthesis.
}
\label{fig:segmentation}
\end{figure}
\subsubsection*{Mitosis segmentation on \dataflywing:}
We now apply \tap on a pixel-level downstream task to fully exploit that the learned \tap representations are dense.
We use the same dataset as for  \dataflywing mitosis classification, but now densely label post-mitotic cells.
We predict a pixel-wise probability map, threshold it at 0.5 and extract connected components as objects.
To evaluate performance, we match a predicted/ground truth object if their intersection over union (IoU) is greater than 0.5, and report the F1 score after matching.
The baseline model is a \unet trained from scratch.
Training a \unet on fixed \tap representations always outperforms the baseline, and 
when only using 3\% of the training data it reaches similar performance as the baseline trained on all available labels (0.67 vs. 0.68,~\cref{fig:segmentation}a).
Interestingly, fine-tuning \tap only slightly outperforms the supervised baseline for this task even for moderate amounts of training data, suggesting that fixed \tap representations generalize better for limited-size datasets.

\subsubsection*{Emerging bud detection on \datayeast:}
Finally, we test \tap on the challenging task of segmenting emerging buds in phase contrast images of yeast colonies. We train \tap networks on \datayeast and generate a dataset of 1205 crops of size $5 \times 192 \times 192$ where we densely label yeast buds in the central frame (defined as buds that appeared less than 13 frames ago) based on available segmentation data~\cite{padovani_cell-acdc_2022}.
We evaluate all methods on held out test videos by interpreting the resulting 2D+time segmentations as 3D objects and computing the F1 score using an IoU threshold of 0.25.
The baseline model is again a \unet trained from scratch.
Surprisingly, training with fixed \tap representations performs slightly worse than the baseline for this dataset (\cref{fig:segmentation}b), possibly due to cell density differencess between \tap training and test videos.
However, fine-tuning \tap features outperforms the baseline by a large margin (\eg 0.64 \vs 0.39 for 120 frames) across the full training data regime, yielding already with 15\% labels the same F1 score as the baseline using all labels.

\section{Discussion}

We have presented \tap, a self-supervised pretraining scheme that learns biologically meaningful representations from live-cell microscopy videos. We show that \tap uncovers sparse time-asymmetric biological processes and events in raw unlabeled recordings without any human supervision.
Furthermore, we demonstrate on a variety of datasets that the learned features can substantially reduce the required amount of annotations for downstream tasks.
Although in this work we focus on 2D+t image sequences, the principle of \tap should generalize to 3D+t datasets, for which dense ground truth creation is often prohibitively expensive and therefore the benefits of modern deep learning are not fully tapped into. We leave this to future work, together with the application of \tap to cell tracking algorithms, in which accurate mitosis detection is a crucial component.

\subsubsection{Acknowledgements}

We thank Albert Dominguez (EPFL) and Uwe Schmidt for helpful comments, Natalie Dye (PoL Dresden) and Franz Gruber for providing the \dataflywing dataset, Benedikt Mairhörmann and Kurt Schmoller (Helmholtz Munich) for providing additional \datayeast training data, and Alan Lowe (UCL) for providing the \dataalan dataset. M.W. and B.G. are supported by the EPFL School of Life Sciences ELISIR program and CARIGEST SA.

\bibliographystyle{splncs04} %

\end{document}